\documentclass[10pt,conference]{IEEEtran}
\IEEEoverridecommandlockouts

\usepackage{cite}
\usepackage{amsmath,amssymb,amsfonts}
\usepackage{algorithmic}
\usepackage{graphicx}
\usepackage{textcomp}
\usepackage{xcolor}
\usepackage{booktabs}
\usepackage{multirow}
\usepackage{adjustbox}
\usepackage{url}

\def\BibTeX{{\rm B\kern-.05em{\sc i\kern-.025em b}\kern-.08em
    T\kern-.1667em\lower.7ex\hbox{E}\kern-.125emX}}

\begin{document}

\title{Engram Memory Network: Brain-Inspired Prototype Explanations\thanks{Presented at IJCNN 2026 (WCCI), Maastricht, Netherlands.}}

\author{
\IEEEauthorblockN{Hyunjun Kim}
\IEEEauthorblockA{
Department of Electrical and Computer Engineering\\
Seoul National University\\
Seoul, South Korea\\
doctor3390@snu.ac.kr
}
\and
\IEEEauthorblockN{Myoung Hoon Ha\textsuperscript{*}}
\IEEEauthorblockA{
Center for Neuroscience-inspired AI\\
KAIST\\
Daejeon, South Korea\\
mh.ha.soar@gmail.com
}
\thanks{*Corresponding author}
}

\maketitle

\begin{abstract}

Artificial intelligence systems are increasingly being deployed in high-stakes decision-making scenarios.
In response, prototype-based models in explainable AI have attracted growing attention for their ability to retrieve visually similar examples as faithful explanations.
Although these models achieve competitive classification accuracy, they often exhibit degraded explanation quality across datasets with varying characteristics, thereby limiting their practical applicability.
To address this limitation, we draw on insights from cognitive neuroscience, particularly the human memory system, which encodes input-relevant information into representations that support recognition.
In this paper, we propose an Engram Memory Network (EMN), a model that emulates the ventral visual stream and inferior temporal cortex in the human brain.
Central to this model is a Hopfield-inspired memory module that iteratively retrieves representative prototypes. These memory-refined latents are used for both reconstruction and classification, encouraging the selected prototypes to capture input-relevant information across diverse datasets.
To address the lack of standardized evaluation for prototype explanations, we assess three complementary aspects: perceptual similarity (LPIPS and DISTS) for perceived resemblance, subclass alignment, measuring whether prototypes match the input's fine-grained category beyond supervised labels, and information preservation (NMI), measuring how much input information is captured by selected prototypes. Experiments show that EMN improves visual similarity by 17.06\%, subclass alignment by 104.57\%, and NMI by 42.42\% on average across CIFAR-10/100, MNIST, HAM10000, and CUB-200, while achieving the highest average classification accuracy (87.62\%) among all methods.
These findings underscore the effectiveness of neuroscience-inspired explanation mechanisms in advancing the interpretability and reliability of prototype-based models.
\end{abstract}


\section{Introduction} \label{sec:int}

Prototype-based explanation models~\cite{Chen2019, DiMarino2025} have emerged as a central approach for explainable AI (XAI; \cite{Narayanan2024}).
They justify model predictions by retrieving examples that resemble the input.
As deep learning extends beyond benchmark classification tasks to high-stakes domains such as medical diagnosis, autonomous driving, and financial decision-making~\cite{Rudin2019}, the demand for accurate and interpretable models has grown. In these applications, prototype-based methods hold particular promise, as they aim to provide explanations that are not only faithful but also intuitive and understandable to human users.
While early methods suffered from limited predictive performance~\cite{Bien2011,Li2018Proto}, recent advances in model architecture and training strategies have substantially improved predictive accuracy~\cite{Xue2024, Ma2024}.
However, consistently achieving high-quality, human-understandable explanations across diverse datasets remains challenging~\cite{nauta2023anecdotal}.

\begin{figure*}[t]
  \centering
  \includegraphics[width=0.95\linewidth]{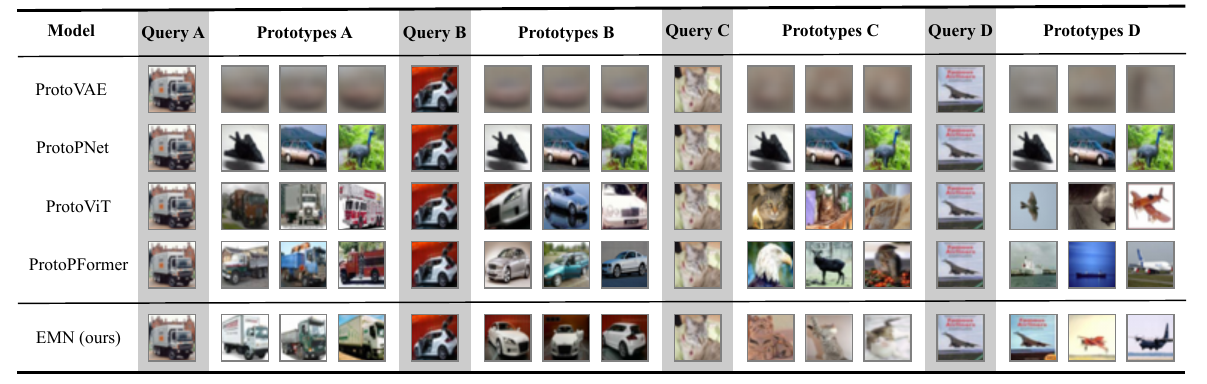}
  \vspace{-0.5em}
  \caption{Queries and selected prototypes by baselines (ProtoVAE~\cite{Gautam2022}, ProtoPNet~\cite{Chen2019}, ProtoViT~\cite{Ma2024}, ProtoPFormer~\cite{Xue2024}) and EMN (ours). Each row shows a model's top-3 retrieved prototypes for four representative queries (A--D); each Query column shows the input image and the three columns on the right shows its retrieved prototypes.
  In these examples, EMN retrieves prototypes that appear visually clearer and more class-consistent, consistent with the improved explanation metrics reported in Sec.~\ref{sec:exp}.
  }
  \vspace{-0.5em}
  \label{fig:1}
\end{figure*}

The explanatory effectiveness of these models varies substantially across datasets. For instance, ProtoVAE~\cite{Gautam2022} reconstructs prototypes from latent vectors and performs well on simple datasets such as MNIST~\cite{lecun2010mnist}, but often produces blurry, ambiguous outputs on more complex datasets such as CIFAR-10~\cite{Krizhevsky2009}. In contrast, patch-based methods such as ProtoPNet~\cite{Chen2019} and ProtoViT~\cite{Ma2024} improve explainability by identifying localized regions of similarity between input and prototypes. These approaches are practical on datasets like CUB-200~\cite{Wah2011}, where object structures are well-defined and redundant across samples. However, when patch-level spatial correspondence between the input and prototypes is weak, as in CIFAR-10~\cite{Krizhevsky2009}, the selected prototypes may not align clearly with the input, ultimately undermining explanatory validity (See Figure~\ref{fig:1}).

In addition, metrics for evaluating whether prototypes are genuinely relevant to inputs remain underexplored~\cite{nauta2023anecdotal}. We propose three complementary metrics: (i) visual similarity, measured by Learned Perceptual Image Patch Similarity (LPIPS~\cite{Zhang2018}) and Deep Image Structure and Texture Similarity (DISTS~\cite{Ding2021}), for human-perceived resemblance; (ii) subclass alignment for fine-grained relevance, evaluated by training on superclass labels and measuring whether prototypes match withheld subclasses---critical in applications like medical diagnosis where within-class distinctions (e.g., cancer subtypes) determine explanation validity; and (iii) information preservation, measured by Normalized Mutual Information (NMI~\cite{cover2006elements}) between input images and their selected prototypes, measuring how much input information prototypes capture.

To address the dataset-dependency problem, we draw on insights from the human visual system, where memory representations are believed to encode input-relevant information for recognition~\cite{liu2021transformative}. 
Accordingly, memories that contribute most to such representations may serve as valid explanations; this motivates our use of prototypes.

In this paper, we propose the Engram Memory Network (EMN), which integrates classification, reconstruction, and explanation. 
Specifically, EMN comprises three components: (i) an encoder that extracts visual features; (ii) an Engram Memory Module (EMM) that performs memory-based association via a modified update rule inspired by a continuous modern Hopfield network~\cite{Ramsauer2021}; and (iii) a decoder that reconstructs the input. Visual and categorical latent vectors are iteratively refined through interaction with prototype memories. After refinement of the latent vectors, the model predicts the label, reconstructs the image, and selects the top-$K$ activated prototypes for explanation. Furthermore, we introduce regularization that promotes prototype diversity and mitigates over-reliance on a narrow subset of prototypes.
We empirically evaluate EMN on diverse benchmarks including CIFAR-10/100~\cite{Krizhevsky2009}, MNIST~\cite{lecun2010mnist}, HAM10000~\cite{tschandl2018ham10000}, and CUB-200~\cite{Wah2011}. On average across datasets, EMN improves visual similarity (LPIPS~\cite{Zhang2018} and DISTS~\cite{Ding2021}) by 17.06\% over ProtoPFormer~\cite{Xue2024} and information preservation (NMI~\cite{cover2006elements}) by 42.42\% over ProtoViT~\cite{Ma2024}. 
EMN achieves 104.57\% higher subclass alignment over ProtoPFormer~\cite{Xue2024}, demonstrating its ability to capture fine-grained semantic structure. For classification, EMN achieves the highest average accuracy (87.62\%) among all compared methods, notably achieving the best performance on HAM10000~\cite{tschandl2018ham10000} (88.51\%).
Our further studies in Sec.~\ref{sec:sensitivity} and \ref{sec:ablation} reveal an explicit accuracy--explanation trade-off: relaxing the separation loss and/or increasing the latent dimensionality improves accuracy (e.g., 97.37\% on CIFAR-10~\cite{Krizhevsky2009}, surpassing most baselines) while moderately reducing explanation quality, which nonetheless remains superior to baseline; we adopt a default setting that prioritizes explanation quality over classification accuracy in this paper.
These results suggest that EMN produces interpretable explanations grounded in memory-based association, achieving consistent performance across diverse datasets~\footnote{
Code is available at \url{https://github.com/attractorlab/engram-prototype}.
}.

Our contributions are threefold:
\begin{itemize}
\item \textit{Brain-Inspired Architecture}: We present a model that structurally reflects biological memory recall, leveraging continuous modern Hopfield networks~\cite{Ramsauer2021} to achieve robust explainability and maintained accuracy across diverse datasets with varying characteristics.
\item \textit{Novel Prototype Evaluation Protocol}: We propose a comprehensive evaluation protocol that assesses visual similarity (LPIPS~\cite{Zhang2018} and DISTS~\cite{Ding2021}), subclass alignment, and NMI~\cite{cover2006elements}.
\item \textit{Cross-dataset Explanatory Robustness}: EMN yields consistently high-quality explanations across diverse datasets, at the cost of a modest decrease in classification accuracy.
\end{itemize}

\section{Related Work} \label{sec:rel}
Post-hoc explanation methods~\cite{Simonyan2014, Selvaraju2017} are widely used in XAI. Among them, attribution-based approaches~\cite{Simonyan2014, Selvaraju2017} generate explanations after prediction by highlighting influential input regions most associated with the output using saliency maps~\cite{Simonyan2014}, attention analysis~\cite{jain2019attention}, and gradient-based attributions~\cite{Selvaraju2017}. However, these explanations are external to the model's decision process, often relying on local cues that may not faithfully reflect the model's internal decision-making process or provide sufficient guidance to users.

In contrast, self-explainable models~\cite{AlvarezMelis2018} embed interpretability mechanisms directly into the model, generating explanations jointly with predictions. Prototype-based models generate explanations by associating each input with prototypical examples, providing case-based justifications grounded in visual evidence. Early CNN-based methods such as ProtoPNet~\cite{Chen2019}, TesNet~\cite{Wang2021TesNet}, and Deformable ProtoPNet~\cite{donnelly2022deformable} rely on patch-to-prototype similarity matching on feature maps, yielding localized explanations. Transformer-based variants like ProtoViT~\cite{Ma2024} and ProtoPFormer~\cite{Xue2024} improve classification accuracy and explanation quality, while ProtoVAE~\cite{Gautam2022} learns class-specific prototypes that can be decoded back into input space for visual verification, and SENN~\cite{AlvarezMelis2018} decomposes predictions into interpretable concept activations and their corresponding relevance scores.

Despite these advances, our experiments indicate that prior prototype-based methods exhibit dataset-specific limitations. Patch-based models struggle when local similarity between inputs and prototypes is weak, as in CIFAR-10~\cite{Krizhevsky2009}, often yielding poorly aligned explanations (See Figure~\ref{fig:1}). Latent-prototype methods like ProtoVAE~\cite{Gautam2022} mitigate this constraint but produce blurry reconstructions, reducing explainability in complex scenarios.

Overall, prior prototype-based methods differ in how they match inputs to prototypes (e.g., patch-level correspondence or latent reconstruction), yet their explanatory effectiveness can vary substantially with dataset characteristics, particularly when spatial correspondence is weak or when reconstructions become ambiguous.

Motivated by these limitations, we propose the EMN. 
Rather than relying solely on similarity-based scoring, EMN grounds prototype selection in memory-refined latent representations that jointly support reconstruction and classification, and selects prototypes with the highest contribution to these latents as explanations. 
By separating visual and categorical latents, EMN enables associative prototype retrieval and yields more consistent explanation quality across datasets while maintaining competitive predictive performance.

\section{Engram Memory Network} \label{sec:mth}

This section presents EMN, which grounds prototype selection in memory-refined latent representations.
The key insight is that prototypes that contribute most to the latents—optimized for both reconstruction and classification—are expected to encode input-relevant information, yielding more consistent explanations across diverse dataset characteristics.
The EMN is designed to emulate the neurocognitive process by which humans recognize objects and retrieve relevant memories for interpretation. Its architecture follows a three-stage structure comprising an encoder, an engram memory module, and a decoder.
We first describe the neurocognitive basis of our approach (Section~\ref{sec:neu}), then detail the encoder (Section~\ref{sec:encoder}), the engram memory module (Section~\ref{sec:emm}), the decoder (Section~\ref{sec:dec}), and the learning objectives (Section~\ref{sec:obj}).

\subsection{Neurocognitive Basis} \label{sec:neu}

\begin{figure*}[t]
  \centering
  \includegraphics[width=0.70\linewidth]{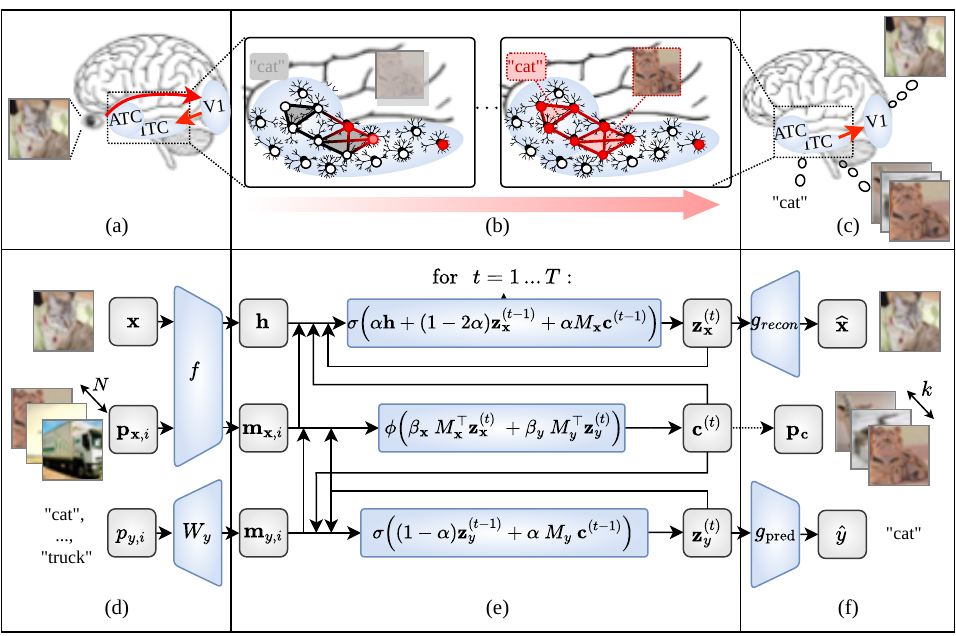}
  \vspace{-0.5em}
  \caption{(a--c) Visual pathway from ventral stream to inferior temporal cortex (ITC), enabling associative memory and categorization with feedback to early visual areas. (d--f) EMN models this process via an encoder, memory-guided refinement, and a decoder.}
  \label{fig:2}
  \vspace{-0.5em}
\end{figure*}

Human object recognition integrates sensory evidence with stored visual knowledge through hierarchical processing in the visual system~\cite{Felleman1991, DiCarlo2012}. The proposed EMN is inspired by this hierarchy and adopts a three-stage architecture, as illustrated in Figure~\ref{fig:2}(a–c).

\textbf{Stage 1: Visual Feature Encoding (Figure~\ref{fig:2}(a)).}
Visual signals propagate from the retina to the primary visual cortex (V1). Along the ventral visual stream,  
representations become progressively more abstract through a hierarchy of visual areas (e.g., V1--V4) and culminate in the inferior temporal cortex (ITC), supporting object recognition~\cite{Hubel1962, DiCarlo2012}. 
EMN models this process using an encoder $f(\cdot)$ that transforms images into visual latent representations (Figure~\ref{fig:2}(d) and Section~\ref{sec:encoder}).

\textbf{Stage 2: Associative Memory Activation (Figure~\ref{fig:2}(b)).}
In the brain, ITC activates associative memories—or engrams—through stimulus-driven and recurrent excitation~\cite{Kar2019}. Formed via Hebbian learning~\cite{Hebb1949}, engrams enable pattern completion: partial input can reactivate the full memory trace~\cite{Josselyn2015}. For example, seeing a cat may evoke related visual or verbal memories.
EMN is motivated by this process using our modified update rule (Section~\ref{sec:emm}), where visual and categorical latent vectors iteratively interact with prototype memories. These interactions integrate three components: continuous input injection, recurrent memory feedback, and state decay, converging to stable representations that encode both perceptual and semantic information (Figure~\ref{fig:2}(e) and Section~\ref{sec:emm}).

\textbf{Stage 3: Recognition and Reconstruction (Figure~\ref{fig:2}(c)).}
Converged activity in ITC informs object classification and memory recall, with feedback signals projecting to early visual areas such as V1. EMN captures this process via a decoder that reconstructs the input from the final latent state (Figure~\ref{fig:2}(f) and Section~\ref{sec:dec}).

Since memory-refined latents are designed to capture sufficient information for both reconstruction and classification, prototypes with high contribution to these latents are likely input-relevant, facilitating valid explanations regardless of dataset characteristics.

\subsection{Encoder} \label{sec:encoder}

As illustrated in Figure~\ref{fig:2}(d), the encoder in EMN transforms visual input into latent representations that serve as the initial substrate for memory interaction and subsequent inference.

\paragraph{Input Encoding}
EMN encodes input images through an encoder $f(\cdot)$, mimicking early visual processing along the ventral stream (Figure~\ref{fig:2}(a)).
Given an input image $\mathbf{x}$, the encoder produces a latent vector $\mathbf{h} = \sigma(f(\mathbf{x})) \in \mathbb{R}^{d_1}$, where $f(\cdot)$ denotes the encoding module.
The activation function $\sigma(\cdot)$ is biologically inspired and defined piecewise as follows:
it outputs $\epsilon x$ for $x < 0$; returns $x$ when $0 \leq x \leq 1$; and computes $\epsilon x + 1 - \epsilon$ for $x > 1$.
This formulation ensures non-zero gradients across the domain while encouraging activations to concentrate near biologically plausible firing thresholds.

\paragraph{Prototype and Engram Initialization}
To model associative memory, EMN initializes a set of $N$ prototype-label pairs $(\mathbf{p}_{\mathbf{x},i}, p_{y,i})$ by randomly sampling from the training set, where $\mathbf{p}_{\mathbf{x},i}$ is the $i$-th prototype image, $p_{y,i}$ is its corresponding class label, and subscripts $\mathbf{x}$ and $y$ denote image and label modalities, respectively.
These are encoded into memory vectors $\mathbf{m}_{\mathbf{x},i} = \sigma(f(\mathbf{p}_{\mathbf{x},i})) \in \mathbb{R}^{d_1}$ and $\mathbf{m}_{y,i} = \sigma(W_y[p_{y,i}]) \in \mathbb{R}^{d_2}$, where $W_y$ is a class embedding matrix. The resulting set $\{\mathbf{m}_{\mathbf{x}, i}, \mathbf{m}_{y, i}\}_{i=1}^N$ forms the engram memory, which supports associative retrieval and pattern completion during inference.

\subsection{Engram Memory Module} \label{sec:emm}

To model associative memory dynamics observed in the brain (Figure~\ref{fig:2}(b) and Section~\ref{sec:neu}), we introduce the EMM as a module for an iterative process that refines latent states by integrating three components: (i) input injection from the encoder, (ii) memory-driven recurrent feedback, and (iii) leaky decay of the current state (Figure~\ref{fig:2}(e)).
Given an encoded input $\mathbf{h} \in \mathbb{R}^{d_1}$, the iterative process produces three outputs after $T$ steps: a refined visual latent vector $\mathbf{z}_\mathbf{x}^{(T)} \in \mathbb{R}^{d_1}$ used for reconstruction, a categorical latent vector $\mathbf{z}_y^{(T)} \in \mathbb{R}^{d_2}$ used for classification, and an activation vector $\mathbf{c}^{(T)} \in \Delta^N$ used to select explanatory prototypes.
At each iteration $t = 1, \dots, T$, the update rules are given by:
\begin{align}
\mathbf{z}_\mathbf{x}^{(0)} &= \mathbf{0},\quad \mathbf{z}_y^{(0)} = \mathbf{0},\quad \mathbf{c}^{(0)} = \frac{\mathbf{1}}{N} \label{eq:init} \\
\mathbf{z}_\mathbf{x}^{(t)} &= \sigma\left( \alpha \mathbf{h} + (1 - 2\alpha) \mathbf{z}_\mathbf{x}^{(t-1)} + \alpha  M_\mathbf{x} \mathbf{c}^{(t-1)}\right) \label{eq:zx} \\
\mathbf{z}_y^{(t)} &= \sigma\left( (1 - \alpha) \mathbf{z}_y^{(t-1)} + \alpha  M_y \mathbf{c}^{(t-1)}\right) \label{eq:zy} \\
\mathbf{c}^{(t)} &= \phi\left( \beta_\mathbf{x} M_\mathbf{x}^\top \mathbf{z}_\mathbf{x}^{(t)} + \beta_y M_y^\top  \mathbf{z}_y^{(t)} \right), \label{eq:c}
\end{align}
where $\sigma(\cdot)$ is the activation function from Section~\ref{sec:encoder}, $\phi(\cdot)$ denotes softmax, $\alpha \in (0, 1)$ controls the update rate, and $\beta_\mathbf{x}, \beta_y > 0$ are learnable coefficients.
Here, $M_\mathbf{x} \in \mathbb{R}^{d_1 \times N}$ and $M_y \in \mathbb{R}^{d_2 \times N}$ denote the memory matrices containing latent visual and label prototypes, respectively. The vector $\mathbf{c}^{(t)}$ reflects the activation levels of the $N$ prototypes based on the current latent state.


Each component of the update mechanism has a specific computational role. Equation~\ref{eq:init} zero-initializes the latent states to begin refinement from a neutral point. In Equations~\ref{eq:zx} and~\ref{eq:zy}, the visual latent vector $\mathbf{z}_\mathbf{x}$ and categorical latent vector $\mathbf{z}_y$ are updated, respectively, via: (i) $\alpha \mathbf{h}$ provides continuous input from the encoder (for $\mathbf{z}_\mathbf{x}$ only), coupling encoder input with memory refinement and enabling end-to-end training; (ii) $\alpha M_\mathbf{x} \mathbf{c}^{(t-1)}$ or $\alpha M_y \mathbf{c}^{(t-1)}$ contributes recurrent feedback weighted by memory activation; (iii) $(1-2\alpha)\mathbf{z}_\mathbf{x}^{(t-1)}$ or $(1-\alpha)\mathbf{z}_y^{(t-1)}$ represents state decay---the difference in decay rates reflects the presence or absence of encoder input. Components (ii), (iii), and the softmax in Equation~\ref{eq:c} follow continuous modern Hopfield dynamics~\cite{Ramsauer2021}, while (i) is introduced to apply Hopfield dynamics in latent space.

\textbf{Differences from~\cite{Ramsauer2021}:} While structurally related to continuous modern Hopfield network~\cite{Ramsauer2021}, this update mechanism differs in three key ways: latent states are zero-initialized (Eq.~\ref{eq:init}); bottom-up input $\mathbf{h}$ is injected at each step (Eq.~\ref{eq:zx}); and the contribution weights $\beta_\mathbf{x}, \beta_y$ are learnable rather than fixed.

\subsection{Decoder} \label{sec:dec}

As shown in Figure~\ref{fig:2}(f), EMN generates three task-specific outputs—image reconstruction, classification, and explanation—based on the refined latent states $\mathbf{z}_\mathbf{x}^{(T)}$, $\mathbf{z}_y^{(T)}$, and the engram activation vector $\mathbf{c}^{(T)}$.
Image reconstruction is performed by a decoder $g_{\text{rec}}$, which maps $\mathbf{z}_\mathbf{x}^{(T)}$ to $\hat{\mathbf{x}} = g_{\text{rec}}(\mathbf{z}_\mathbf{x}^{(T)})$, reflecting feedback from ITC to early visual areas. The decoder architecture is flexible and task-agnostic.
Classification uses a linear head $g_{\text{pred}}$ applied to $\mathbf{z}_y^{(T)}$:
$\hat{\mathbf{y}} = \beta_y \sigma(W_y) \mathbf{z}_y^{(T)}$. This design reflects the alignment between categorical activation and class prediction.
Explanations are derived from $\mathbf{c}^{(T)}$ by selecting the top-$K$ activated prototypes, emulating memory-based associative retrieval.
These decoding pathways instantiate perception, recognition, and memory retrieval, reflecting distinct cognitive functional roles.
Because the same memory-refined latent supports both reconstruction and classification, prototypes that contribute most to this latent are likely to encode input-relevant information, providing a principled basis for explanation validity across diverse domains.

\subsection{Learning Objectives} \label{sec:obj}

EMN is trained using a multi-objective loss that jointly optimizes classification, reconstruction, and explanation quality:
\[
\mathcal{L}_{\text{total}} = \lambda_{\text{cls}} \mathcal{L}_{\text{cls}} + \lambda_{\text{rec}} \mathcal{L}_{\text{rec}} + \lambda_{\text{div}} \mathcal{L}_{\text{div}} + \lambda_{\text{sep}} \mathcal{L}_{\text{sep}},
\]
where each $\lambda$ balances the contribution of the respective term.
The classification loss ensures that the label latent vector predicts the correct class:
$
\mathcal{L}_{\text{cls}} = \text{CrossEntropy}(\hat{\mathbf{y}}, y),
$
encouraging class-consistent engram activations. To retain visual fidelity, the reconstruction loss minimizes the discrepancy between the input and reconstructed image:
$
\mathcal{L}_{\text{rec}} = \|\mathbf{x} - \hat{\mathbf{x}}\|_2^2.
$
To mitigate prototype overuse and collapse (see Figure~\ref{fig:1}), EMN introduces two entropy-based losses that regulate memory utilization.
The diversity loss promotes global usage of the memory space by maximizing entropy over the batch-averaged activation distribution:
$
\mathcal{L}_{\text{div}} = -\mathcal{H}(\bar{\mathbf{c}}) = \sum_{i=1}^N \bar{c}_i \log \bar{c}_i,
$
where $\bar{\mathbf{c}} = \frac{1}{B} \sum_{j=1}^B \mathbf{c}^{(T)}_j$.
The separation loss encourages local selectivity by aligning each sample's prototype entropy to a target:
$
\mathcal{L}_{\text{sep}} = \left| \frac{1}{B} \sum_{j=1}^B \mathcal{H}(\mathbf{c}_j^{(T)}) - \log K \right|.
$
This guides each explanation to rely on a small number of relevant prototypes—typically around three—to balance richness and focus.
Together, $\mathcal{L}_{\text{div}}$ and $\mathcal{L}_{\text{sep}}$ shape the softmax-driven memory dynamics to ensure explanations are diverse across inputs and selective within each instance, enabling EMN to mimic human-like associative recall.

\section{Experiments} \label{sec:exp}

We validate the effectiveness of the proposed EMN through comprehensive experiments.

\textbf{Datasets.} We evaluate on CIFAR-10~\cite{Krizhevsky2009}, CIFAR-100~\cite{Krizhevsky2009}, MNIST~\cite{lecun2010mnist}, HAM10000~\cite{tschandl2018ham10000} (medical dermatology), and CUB-200~\cite{Wah2011} (fine-grained bird classification).

\textbf{Model Architecture.} Although EMN is compatible with various backbone architectures, we adopt Vision Transformer (ViT)~\cite{dosovitskiy2021image} for all primary comparisons to maintain consistency with recent prototype-based models. The encoder uses DeiT-Small~\cite{pmlr-v139-touvron21a}\footnote{timm library: \url{https://github.com/huggingface/pytorch-image-models}} and the decoder uses a modified ViT architecture.

\textbf{Training Process.} Training proceeds in three stages: (1) the encoder is initialized with DeiT-Small weights pretrained on ImageNet~\cite{deng2009imagenet}; (2) the encoder and decoder are jointly trained on the target dataset using classification and reconstruction losses without the engram memory module, for 30 epochs on CIFAR and 50 epochs on other datasets; (3) the full EMN is then trained for 50 epochs on all datasets.

\textbf{Hyperparameters.} We use $N{=}2000$ prototypes randomly sampled from the training set with $K{=}3$ for explanation. Latent dimensions are $d_1{=}4096$ (visual) and $d_2{=}512$ (categorical). Memory refinement runs for $T{=}24$ iterations with update rate $\alpha{=}0.05$ and $\varepsilon{=}0.001$ in $\sigma(\cdot)$. Loss weights are $\lambda_{\text{cls}}{=}\lambda_{\text{rec}}{=}\lambda_{\text{div}}{=}\lambda_{\text{sep}}{=}1$. Training uses AdamW~\cite{loshchilov2019decoupled} with lr=$10^{-5}$, weight decay 0.05, batch size 64. Data augmentation includes RandomHorizontalFlip and ColorJitter (brightness=0.2, contrast=0.2, saturation=0.2, and hue=0.1).

\textbf{Baselines.} We compare EMN against ProtoVAE~\cite{Gautam2022}, ProtoPNet~\cite{Chen2019}, ProtoViT~\cite{Ma2024}, and ProtoPFormer~\cite{Xue2024}. All baselines are trained and evaluated according to their official protocols, using the same data splits, preprocessing, and augmentation. All models are implemented in PyTorch~\cite{paszke2019pytorch}.

\textbf{Experimental Environment.} All experiments were conducted on a single NVIDIA GPU (32GB VRAM) running Ubuntu Linux. Training EMN on CIFAR-10 takes approximately 3 hours across all training stages.

\subsection{Visual Similarity}
\label{sec:vis-sim}

\begin{table}[t]
\centering
\caption{
Visual similarity (LPIPS~\cite{Zhang2018} and DISTS~\cite{Ding2021}) comparison across datasets. Results are reported as mean $\pm$ std. ($\times 10^{-1}$; lower is better; bold indicates best).
}
\vspace{-0.5em} 
\label{tab:vis-sim}
\resizebox{\columnwidth}{!}{%
\begin{tabular}{l@{\hskip 4pt}c@{\hskip 4pt}c@{\hskip 4pt}c@{\hskip 4pt}c}
\toprule
Model &
\multicolumn{2}{c}{LPIPS~\cite{Zhang2018}} &
\multicolumn{2}{c}{DISTS~\cite{Ding2021}} \\
& top1 & top3 & top1 & top3 \\
\midrule
\multicolumn{5}{l}{\textbf{(CIFAR-10~\cite{Krizhevsky2009})}} \\
ProtoVAE~\cite{Gautam2022}      & 7.50 $\pm$ 0.06 & 7.50 $\pm$ 0.07 & 6.75 $\pm$ 0.11 & 6.76 $\pm$ 0.11 \\
ProtoPNet~\cite{Chen2019}      & 6.95 $\pm$ 0.19 & 7.00 $\pm$ 0.32 & 4.07 $\pm$ 0.13 & 4.02 $\pm$ 0.12 \\
ProtoViT~\cite{Ma2024}       & 6.26 $\pm$ 0.10 & 6.31 $\pm$ 0.07 & 3.67 $\pm$ 0.03 & 3.69 $\pm$ 0.03 \\
ProtoPFormer~\cite{Xue2024}  & 6.23 $\pm$ 0.02 & 6.24 $\pm$ 0.02 & 3.68 $\pm$ 0.01 & 3.69 $\pm$ 0.02 \\
EMN (ours)    & \textbf{5.38 $\pm$ 0.02} & \textbf{5.50 $\pm$ 0.02} & \textbf{3.22 $\pm$ 0.01} & \textbf{3.28 $\pm$ 0.01} \\
\addlinespace
\multicolumn{5}{l}{\textbf{(CIFAR-100~\cite{Krizhevsky2009})}} \\
ProtoVAE~\cite{Gautam2022}      & 7.27 $\pm$ 0.06 & 7.27 $\pm$ 0.06 & 6.32 $\pm$ 0.07 & 6.32 $\pm$ 0.07 \\
ProtoPNet~\cite{Chen2019}      & 6.49 $\pm$ 0.08 & 6.57 $\pm$ 0.08 & 3.83 $\pm$ 0.04 & 3.88 $\pm$ 0.03 \\
ProtoViT~\cite{Ma2024}       & 6.11 $\pm$ 0.02 & 6.22 $\pm$ 0.01 & 3.57 $\pm$ 0.01 & 3.62 $\pm$ 0.00 \\
ProtoPFormer~\cite{Xue2024}  & 6.34 $\pm$ 0.04 & 6.35 $\pm$ 0.02 & 3.71 $\pm$ 0.02 & 3.71 $\pm$ 0.02 \\
EMN (ours)    & \textbf{5.66 $\pm$ 0.02} & \textbf{5.85 $\pm$ 0.01} & \textbf{3.38 $\pm$ 0.01} & \textbf{3.49 $\pm$ 0.01} \\
\addlinespace
\multicolumn{5}{l}{\textbf{(MNIST~\cite{lecun2010mnist})}} \\
ProtoVAE~\cite{Gautam2022}      & 2.33 $\pm$ 0.06 & 2.34 $\pm$ 0.06 & 2.38 $\pm$ 0.02 & 2.38 $\pm$ 0.02 \\
ProtoPNet~\cite{Chen2019}      & 6.71 $\pm$ 0.05 & 6.82 $\pm$ 0.08 & 4.28 $\pm$ 0.05 & 4.37 $\pm$ 0.07 \\
ProtoViT~\cite{Ma2024}       & 6.59 $\pm$ 0.06 & 6.73 $\pm$ 0.07 & 4.12 $\pm$ 0.02 & 4.20 $\pm$ 0.03 \\
ProtoPFormer~\cite{Xue2024}  & 3.06 $\pm$ 0.05 & 3.05 $\pm$ 0.03 & 2.22 $\pm$ 0.01 & 2.22 $\pm$ 0.01 \\
EMN (ours)    & \textbf{1.70 $\pm$ 0.01} & \textbf{1.82 $\pm$ 0.01} & \textbf{1.73 $\pm$ 0.01} & \textbf{1.78 $\pm$ 0.01} \\
\addlinespace
\multicolumn{5}{l}{\textbf{(HAM10000~\cite{tschandl2018ham10000})}} \\
ProtoVAE~\cite{Gautam2022}      & -- & -- & -- & -- \\
ProtoPNet~\cite{Chen2019}      & 5.19 $\pm$ 0.23 & 5.35 $\pm$ 0.21 & 3.43 $\pm$ 0.08 & 3.48 $\pm$ 0.07 \\
ProtoViT~\cite{Ma2024}       & 4.69 $\pm$ 0.17 & 4.71 $\pm$ 0.16 & 3.15 $\pm$ 0.03 & 3.20 $\pm$ 0.03 \\
ProtoPFormer~\cite{Xue2024}  & 5.40 $\pm$ 0.53 & 5.40 $\pm$ 0.47 & 3.61 $\pm$ 0.44 & 3.64 $\pm$ 0.38 \\
EMN (ours)    & \textbf{3.50 $\pm$ 0.02} & \textbf{3.59 $\pm$ 0.01} & \textbf{2.40 $\pm$ 0.02} & \textbf{2.46 $\pm$ 0.02} \\
\addlinespace
\multicolumn{5}{l}{\textbf{(CUB-200~\cite{Wah2011})}} \\
ProtoVAE~\cite{Gautam2022}      & -- & -- & -- & -- \\
ProtoPNet~\cite{Chen2019}      & 6.79 $\pm$ 0.05 & 6.81 $\pm$ 0.05 & 4.12 $\pm$ 0.03 & 4.13 $\pm$ 0.02 \\
ProtoViT~\cite{Ma2024}       & 6.82 $\pm$ 0.00 & 6.87 $\pm$ 0.00 & 3.97 $\pm$ 0.00 & 4.00 $\pm$ 0.00 \\
ProtoPFormer~\cite{Xue2024}  & 6.77 $\pm$ 0.03 & 6.80 $\pm$ 0.02 & 3.72 $\pm$ 0.02 & 3.75 $\pm$ 0.02 \\
EMN (ours)    & \textbf{6.44 $\pm$ 0.02} & \textbf{6.55 $\pm$ 0.01} & \textbf{3.55 $\pm$ 0.01} & \textbf{3.62 $\pm$ 0.00} \\
\bottomrule
\end{tabular}%
}
\end{table}

This section assesses the visual similarity between input queries and selected prototypes. Since explanations are intended for human users, the ideal evaluation would rely on human judgments; however, due to feasibility constraints, we employ proxy metrics---LPIPS~\cite{Zhang2018} and DISTS~\cite{Ding2021}---that align with human perceptual judgments, while imperfect but scalable for large-scale evaluation.

As shown in Table~\ref{tab:vis-sim}, EMN achieves the best visual similarity (lowest LPIPS~\cite{Zhang2018} and DISTS~\cite{Ding2021}) across all five datasets, with an average improvement of 17.06\% over the best baseline (ProtoPFormer~\cite{Xue2024}). These results indicate that EMN retrieves prototypes more perceptually aligned with input queries, providing more interpretable explanations as illustrated in Figure~\ref{fig:1}.
ProtoVAE~\cite{Gautam2022} is excluded from CUB-200~\cite{Wah2011} and HAM10000~\cite{tschandl2018ham10000} as its decoded prototypes become uninformative and blurry on these datasets.

\subsection{Subclass Alignment} \label{sec:subclass}

\begin{table}[t]
\centering
\caption{Subclass Alignment (\%) on CIFAR-100S (higher is better; chance level is 1\%).} \label{tab:subclass}
\vspace{-0.5em}
\begin{tabular}{l@{\hskip 4pt}c@{\hskip 4pt}c}
\toprule
\textbf{Method} & \textbf{top1} & \textbf{top3} \\
\midrule
ProtoVAE~\cite{Gautam2022}     & -        & -        \\
ProtoPNet~\cite{Chen2019}     & 2.28 $\pm$ 0.91  & 1.95 $\pm$ 0.47  \\
ProtoViT~\cite{Ma2024}     & 15.93 $\pm$ 6.55 & 14.03 $\pm$ 7.04 \\
ProtoPFormer~\cite{Xue2024} & 19.49 $\pm$ 1.60 & 19.42 $\pm$ 1.38 \\
EMN (ours)   & \textbf{39.87 $\pm$ 0.72} & \textbf{36.07 $\pm$ 0.36} \\
\bottomrule
\end{tabular}
\end{table}

Representation learning aims to preserve the latent geometry of real-world objects so that semantically similar instances are placed in close proximity. Since latent structure is unobservable, we use hierarchical labels (superclass--subclass) as a proxy: e.g., ``Pine'' and ``Oak'' within ``Tree'' share more features than items from other superclasses. Models are trained using only superclass labels while subclass labels are withheld; at evaluation, we reveal subclass labels as hidden ground truth to measure prototype--query alignment.
To evaluate whether models identify prototypes that are class-consistent and visually aligned with the query at a finer subclass level, we conduct a hierarchical matching experiment using CIFAR-100S, which reorganizes CIFAR-100~\cite{Krizhevsky2009}'s 100 classes into 20 superclasses (5 subclasses each).\footnote{\url{https://github.com/BillyBSig/CIFAR-100-TFDS}} ProtoVAE~\cite{Gautam2022} is excluded as its generated prototypes lack subclass labels. Results are summarized in Table~\ref{tab:subclass}.
Table~\ref{tab:subclass} shows that EMN achieves 39.87\% (top-1) and 36.07\% (top-3) subclass alignment, 104.57\% higher than the best baseline ProtoPFormer~\cite{Xue2024} (19.49\%, 19.42\%). This substantial improvement demonstrates that EMN effectively captures fine-grained visual features, enabling input-specific prototype selection rather than generic class-level matching.

\subsection{Classification}

\begin{table}[t]
\centering
\caption{Classification accuracy (\%) across all datasets (higher is better; bold indicates best).} \label{tab:cls}
\begin{tabular}{lclc}
\toprule
Model & Acc. & Model & Acc. \\
\midrule
\multicolumn{2}{l}{\textbf{(CIFAR-10~\cite{Krizhevsky2009})}} & \multicolumn{2}{l}{\textbf{(MNIST~\cite{lecun2010mnist})}} \\
ProtoVAE~\cite{Gautam2022} & 81.59 $\pm$ 0.36 & ProtoVAE~\cite{Gautam2022} & 99.05 $\pm$ 0.02 \\
ProtoPNet~\cite{Chen2019} & 94.30 $\pm$ 0.12 & ProtoPNet~\cite{Chen2019} & 99.06 $\pm$ 0.10 \\
ProtoViT~\cite{Ma2024} & 97.63 $\pm$ 0.08 & ProtoViT~\cite{Ma2024} & 99.09 $\pm$ 0.09 \\
ProtoPFormer~\cite{Xue2024} & \textbf{97.85 $\pm$ 0.06} & ProtoPFormer~\cite{Xue2024} & \textbf{99.41 $\pm$ 0.05} \\
EMN (ours) & 96.39 $\pm$ 0.27 & EMN (ours) & 99.19 $\pm$ 0.08 \\
\addlinespace
\multicolumn{2}{l}{\textbf{(CIFAR-100~\cite{Krizhevsky2009})}} & \multicolumn{2}{l}{\textbf{(HAM10000~\cite{tschandl2018ham10000})}} \\
ProtoVAE~\cite{Gautam2022} & 50.37 $\pm$ 0.89 & ProtoVAE~\cite{Gautam2022} & -- \\
ProtoPNet~\cite{Chen2019} & 68.63 $\pm$ 0.35 & ProtoPNet~\cite{Chen2019} & 62.92 $\pm$ 2.84 \\
ProtoViT~\cite{Ma2024} & 86.41 $\pm$ 0.13 & ProtoViT~\cite{Ma2024} & 80.49 $\pm$ 0.68 \\
ProtoPFormer~\cite{Xue2024} & \textbf{87.83 $\pm$ 0.25} & ProtoPFormer~\cite{Xue2024} & 66.95 $\pm$ 0.00 \\
EMN (ours) & 85.28 $\pm$ 0.21 & EMN (ours) & \textbf{88.51 $\pm$ 1.02} \\
\addlinespace
\multicolumn{4}{l}{\textbf{(CUB-200~\cite{Wah2011})}} \\
ProtoVAE~\cite{Gautam2022} & -- & & \\
ProtoPNet~\cite{Chen2019} & 22.43 $\pm$ 2.29 & & \\
ProtoViT~\cite{Ma2024} & \textbf{74.09 $\pm$ 0.24} & & \\
ProtoPFormer~\cite{Xue2024} & 50.51 $\pm$ 1.61 & & \\
EMN (ours) & 68.74 $\pm$ 0.54 & & \\
\bottomrule
\end{tabular}
\end{table}

Table~\ref{tab:cls} reports classification accuracy across all datasets. EMN achieves the highest average accuracy (87.62\%) among all methods, outperforming the best baseline ProtoViT~\cite{Ma2024} (87.54\%). Notably, on HAM10000~\cite{tschandl2018ham10000} (medical imaging), EMN achieves the highest accuracy (88.51\%), outperforming the best baseline by 8 percentage points.

\subsection{Prototype Information Content} \label{sec:nmi}

\begin{table}[t]
\centering
\caption{$\text{NMI}(X;P)$~\cite{cover2006elements} scores across all datasets ($\times 10^{-1}$; higher is better; bold indicates best).} \label{tab:nmi}
\begin{tabular}{l@{\hskip 6pt}c@{\hskip 6pt}c}
\toprule
Model & top1 & top3 \\
\midrule
\multicolumn{3}{l}{\textbf{(CIFAR-10~\cite{Krizhevsky2009})}} \\
ProtoVAE~\cite{Gautam2022}      & 0.07 $\pm$ 0.02 & \textbf{2.41 $\pm$ 1.37} \\
ProtoPNet~\cite{Chen2019}      & 0.07 $\pm$ 0.10 & 0.08 $\pm$ 0.12 \\
ProtoViT~\cite{Ma2024}       & 0.60 $\pm$ 0.08 & 0.50 $\pm$ 0.05 \\
ProtoPFormer~\cite{Xue2024}  & 0.55 $\pm$ 0.05 & 0.55 $\pm$ 0.04 \\
EMN (ours)    & \textbf{1.26 $\pm$ 0.02} & 1.21 $\pm$ 0.02 \\
\addlinespace
\multicolumn{3}{l}{\textbf{(CIFAR-100~\cite{Krizhevsky2009})}} \\
ProtoVAE~\cite{Gautam2022}      & 0.04 $\pm$ 0.01 & 0.06 $\pm$ 0.08 \\
ProtoPNet~\cite{Chen2019}      & 0.29 $\pm$ 0.01 & 0.24 $\pm$ 0.01 \\
ProtoViT~\cite{Ma2024}       & 0.49 $\pm$ 0.03 & 0.44 $\pm$ 0.01 \\
ProtoPFormer~\cite{Xue2024}  & 0.44 $\pm$ 0.03 & 0.42 $\pm$ 0.02 \\
EMN (ours)    & \textbf{1.05 $\pm$ 0.04} & \textbf{0.97 $\pm$ 0.03} \\
\addlinespace
\multicolumn{3}{l}{\textbf{(MNIST~\cite{lecun2010mnist})}} \\
ProtoVAE~\cite{Gautam2022}      & \textbf{3.08 $\pm$ 0.06} & \textbf{3.05 $\pm$ 0.06} \\
ProtoPNet~\cite{Chen2019}      & -- & -- \\
ProtoViT~\cite{Ma2024}       & 1.84 $\pm$ 0.10 & 1.35 $\pm$ 0.10 \\
ProtoPFormer~\cite{Xue2024}  & 1.78 $\pm$ 0.10 & 1.73 $\pm$ 0.08 \\
EMN (ours)    & 2.62 $\pm$ 0.05 & 2.49 $\pm$ 0.03 \\
\addlinespace
\multicolumn{3}{l}{\textbf{(HAM10000~\cite{tschandl2018ham10000})}} \\
ProtoVAE~\cite{Gautam2022}      & -- & -- \\
ProtoPNet~\cite{Chen2019}      & 0.68 $\pm$ 0.38 & 0.63 $\pm$ 0.35 \\
ProtoViT~\cite{Ma2024}       & 0.97 $\pm$ 0.15 & 0.78 $\pm$ 0.10 \\
ProtoPFormer~\cite{Xue2024}  & 0.00 $\pm$ 0.00 & 0.00 $\pm$ 0.00 \\
EMN (ours)    & \textbf{1.18 $\pm$ 0.04} & \textbf{1.09 $\pm$ 0.04} \\
\addlinespace
\multicolumn{3}{l}{\textbf{(CUB-200~\cite{Wah2011})}} \\
ProtoVAE~\cite{Gautam2022}      & -- & -- \\
ProtoPNet~\cite{Chen2019}      & 0.70 $\pm$ 0.03 & 0.65 $\pm$ 0.03 \\
ProtoViT~\cite{Ma2024}       & 1.38 $\pm$ 0.01 & 1.33 $\pm$ 0.01 \\
ProtoPFormer~\cite{Xue2024}  & 1.38 $\pm$ 0.03 & 1.32 $\pm$ 0.02 \\
EMN (ours)    & \textbf{1.41 $\pm$ 0.03} & \textbf{1.34 $\pm$ 0.02} \\
\bottomrule
\end{tabular}
\end{table}

To evaluate whether prototypes capture meaningful information, we compute $\text{NMI}(X;P)$~\cite{cover2006elements}\footnote{Calculated using k-nearest-neighbor--based entropy estimation from the NPEET package: \url{https://github.com/gregversteeg/NPEET}}, which measures how well prototypes preserve input-specific visual details. As shown in Table~\ref{tab:nmi}, EMN achieves the best $\text{NMI}(X;P)$~\cite{cover2006elements} on four of five datasets, with an improvement of 42.42\% over the best baseline (ProtoViT~\cite{Ma2024}). The exception is MNIST~\cite{lecun2010mnist}, where ProtoVAE~\cite{Gautam2022} achieves higher $\text{NMI}(X;P)$~\cite{cover2006elements} due to its reconstruction-based approach being well-suited to simple, low-variance images. Overall, these results indicate that EMN's selected prototypes reflect individual query characteristics, aligning with its design goal of retrieving visually similar examples rather than generic class representatives.

\begin{figure*}[!t]
\centering
\includegraphics[width=\textwidth]{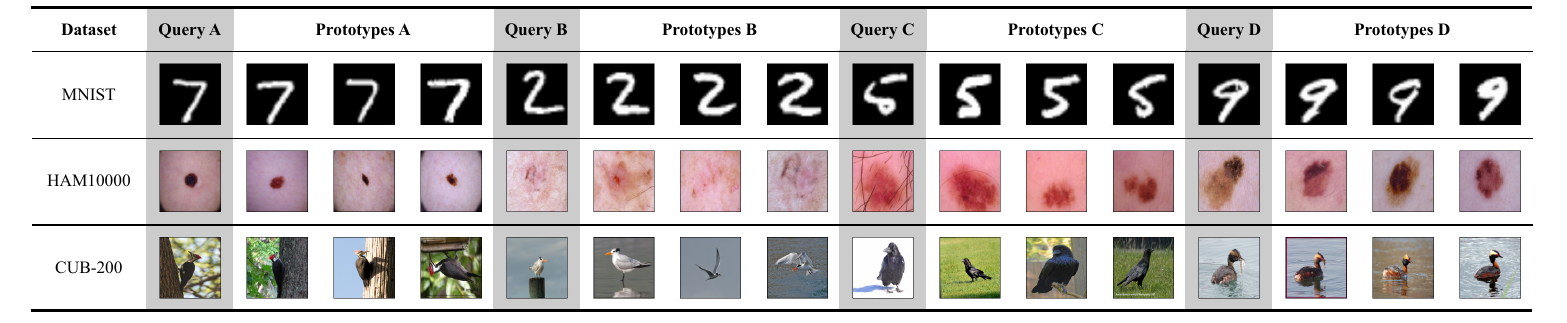}
\vspace{-0.5em}
\caption{Examples of prototype-based explanations across datasets: (from top to bottom) 
MNIST~\cite{lecun2010mnist}, HAM10000~\cite{tschandl2018ham10000}, and CUB-200~\cite{Wah2011}. The Query A-D columns show input queries, and the Prototype A-D column groups display the corresponding top-3 retrieved prototypes.}
\label{fig:qualitative}
\vspace{-0.5em}
\end{figure*}

\subsection{Sensitivity Analysis} \label{sec:sensitivity}

\begin{table}[t]
\centering
\caption{Sensitivity analysis on CIFAR-10~\cite{Krizhevsky2009}. $N$: number of prototypes; $K$: target entropy for $\mathcal{L}_{\text{sep}}$; $d_2$: categorical latent dimension. Acc (\%) higher is better; LPIPS~\cite{Zhang2018}, DISTS~\cite{Ding2021} ($\times 10^{-1}$) lower is better; NMI~\cite{cover2006elements} ($\times 10^{-1}$) higher is better.}
\vspace{-0.5em}
\label{tab:sensitivity}
\resizebox{\columnwidth}{!}{%
\begin{tabular}{lcccc}
\toprule
\textbf{Setting} & \textbf{Acc$\uparrow$} & \textbf{LPIPS~\cite{Zhang2018}$\downarrow$} & \textbf{DISTS~\cite{Ding2021}$\downarrow$} & \textbf{NMI$(X;P)$~\cite{cover2006elements}$\uparrow$} \\
\midrule
$N{=}200$ & 96.63 $\pm$ 0.20 & 5.75 $\pm$ 0.02 & 3.39 $\pm$ 0.01 & 1.06 $\pm$ 0.04 \\
$N{=}500$ & 96.63 $\pm$ 0.30 & 5.59 $\pm$ 0.02 & 3.30 $\pm$ 0.01 & 1.17 $\pm$ 0.02 \\
$N{=}2000$ & 96.69 $\pm$ 0.25 & 5.40 $\pm$ 0.02 & 3.22 $\pm$ 0.01 & 1.21 $\pm$ 0.03 \\
$N{=}5000$ & 96.69 $\pm$ 0.19 & \textbf{5.34 $\pm$ 0.03} & \textbf{3.18 $\pm$ 0.02} & \textbf{1.28 $\pm$ 0.02} \\
\midrule
$K{=}1$ & 96.17 $\pm$ 0.24 & 5.50 $\pm$ 0.04 & 3.33 $\pm$ 0.02 & \textbf{1.24 $\pm$ 0.02} \\
$K{=}3$ & 96.69 $\pm$ 0.25 & \textbf{5.40 $\pm$ 0.02} & \textbf{3.22 $\pm$ 0.01} & 1.21 $\pm$ 0.03 \\
$K{=}5$ & 96.72 $\pm$ 0.12 & 5.41 $\pm$ 0.03 & \textbf{3.22 $\pm$ 0.01} & 1.21 $\pm$ 0.04 \\
$K{=}7$ & \textbf{96.73 $\pm$ 0.25} & 5.41 $\pm$ 0.01 & \textbf{3.22 $\pm$ 0.01} & 1.21 $\pm$ 0.03 \\
\midrule
$d_2{=}128$ & 96.57 $\pm$ 0.26 & \textbf{5.38 $\pm$ 0.03} & \textbf{3.21 $\pm$ 0.01} & \textbf{1.25 $\pm$ 0.03} \\
$d_2{=}256$ & 96.62 $\pm$ 0.25 & 5.40 $\pm$ 0.01 & \textbf{3.21 $\pm$ 0.01} & 1.22 $\pm$ 0.02 \\
$d_2{=}512$ & 96.69 $\pm$ 0.25 & 5.40 $\pm$ 0.02 & 3.22 $\pm$ 0.01 & 1.21 $\pm$ 0.03 \\
$d_2{=}1024$ & \textbf{96.72 $\pm$ 0.21} & 5.41 $\pm$ 0.03 & 3.22 $\pm$ 0.01 & 1.24 $\pm$ 0.04 \\
\bottomrule
\end{tabular}%
}
\end{table}

Table~\ref{tab:sensitivity} shows sensitivity analysis on CIFAR-10~\cite{Krizhevsky2009}. Increasing the number of prototypes $N$ (from 200 to 5000) consistently improves explanation quality: LPIPS~\cite{Zhang2018} ($5.75{\rightarrow}5.34$), DISTS~\cite{Ding2021} ($3.39{\rightarrow}3.18$), and NMI$(X;P)$~\cite{cover2006elements} ($1.06{\rightarrow}1.28$), while classification accuracy remains stable ($96.63$--$96.69$\%). This indicates that larger memory provides richer retrieval candidates. The target entropy $K$ in $\mathcal{L}_{\text{sep}}$ shows robust behavior across $K{\in}\{1,3,5,7\}$: accuracy $96.17$--$96.73$\%, LPIPS~\cite{Zhang2018} $5.40$--$5.50$, DISTS~\cite{Ding2021} $3.22$--$3.33$, and NMI$(X;P)$~\cite{cover2006elements} $1.21$--$1.24$. The categorical latent dimension $d_2$ reveals a trade-off: as $d_2$ increases from 128 to 1024, LPIPS~\cite{Zhang2018} increases ($5.38{\rightarrow}5.41$), DISTS~\cite{Ding2021} remains stable ($3.21{\rightarrow}3.22$), and NMI$(X;P)$~\cite{cover2006elements} slightly decreases ($1.25{\rightarrow}1.24$), while accuracy slightly improves ($96.57{\rightarrow}96.72$\%). These results confirm that EMN is robust to hyperparameter choices, with $d_2$ being the only source of a modest accuracy--explanation trade-off. We adopt $N{=}2000$, $K{=}3$, and $d_2{=}512$ as defaults.

\subsection{Ablation Study} \label{sec:ablation}

\begin{table}[t]
\centering
\caption{Ablation study on loss components (CIFAR-10~\cite{Krizhevsky2009}). Acc (\%) higher is better; LPIPS~\cite{Zhang2018}, DISTS~\cite{Ding2021} ($\times 10^{-1}$) lower is better; NMI~\cite{cover2006elements} ($\times 10^{-1}$) higher is better.}
\vspace{-0.5em}
\label{tab:ablation}
\resizebox{\columnwidth}{!}{%
\begin{tabular}{lcccc}
\toprule
\textbf{Setting} & \textbf{Acc$\uparrow$} & \textbf{LPIPS~\cite{Zhang2018}$\downarrow$} & \textbf{DISTS~\cite{Ding2021}$\downarrow$} & \textbf{NMI$(X;P)$~\cite{cover2006elements}$\uparrow$} \\
\midrule
EMN (ours) & 96.69 $\pm$ 0.25 & 5.40 $\pm$ 0.02 & \textbf{3.22 $\pm$ 0.01} & 1.21 $\pm$ 0.03 \\
w/o $\mathcal{L}_{\text{rec}}$ & 96.67 $\pm$ 0.28 & \textbf{5.39 $\pm$ 0.02} & \textbf{3.22 $\pm$ 0.01} & 1.24 $\pm$ 0.02 \\
w/o $\mathcal{L}_{\text{div}}$ & 96.44 $\pm$ 0.33 & 5.43 $\pm$ 0.03 & 3.31 $\pm$ 0.02 & 1.27 $\pm$ 0.04 \\
w/o $\mathcal{L}_{\text{sep}}$ & 97.37 $\pm$ 0.17 & 5.48 $\pm$ 0.02 & 3.26 $\pm$ 0.01 & 1.17 $\pm$ 0.01 \\
w/o $\mathcal{L}_{\text{rec}}, \mathcal{L}_{\text{div}}$ & 96.36 $\pm$ 0.19 & 5.41 $\pm$ 0.04 & 3.30 $\pm$ 0.02 & 1.27 $\pm$ 0.05 \\
w/o $\mathcal{L}_{\text{rec}}, \mathcal{L}_{\text{sep}}$ & \textbf{97.38 $\pm$ 0.15} & 5.57 $\pm$ 0.02 & 3.32 $\pm$ 0.00 & 1.16 $\pm$ 0.02 \\
w/o $\mathcal{L}_{\text{div}}, \mathcal{L}_{\text{sep}}$ & 96.74 $\pm$ 0.55 & 5.82 $\pm$ 0.12 & 3.60 $\pm$ 0.13 & \textbf{1.49 $\pm$ 0.20} \\
w/o all & 97.15 $\pm$ 0.17 & 5.54 $\pm$ 0.04 & 3.43 $\pm$ 0.02 & 1.31 $\pm$ 0.05 \\
\bottomrule
\end{tabular}%
}
\end{table}

Table~\ref{tab:ablation} examines the contribution of each loss component. Individual removal of $\mathcal{L}_{\text{div}}$ or $\mathcal{L}_{\text{sep}}$ produces only modest changes, but removing both substantially degrades visual similarity (LPIPS~\cite{Zhang2018} $5.40{\rightarrow}5.82$, DISTS~\cite{Ding2021} $3.22{\rightarrow}3.60$), demonstrating their synergistic effect. Notably, removing $\mathcal{L}_{\text{sep}}$ alone improves accuracy to $97.37$\%, revealing an accuracy--explanation trade-off. Nevertheless, even under this setting, EMN still outperforms the best baseline in visual similarity (e.g., LPIPS~\cite{Zhang2018} $5.82$ vs.\ $6.23$ for ProtoPFormer~\cite{Xue2024}). 
These results reveal that $\mathcal{L}_{\text{sep}}$ controls the accuracy--explanation trade-off, while both losses together are essential for visual similarity. We adopt the full configuration to prioritize explanation quality.

\subsection{Examples of Prototype-based Explanations}

Figure~\ref{fig:qualitative} presents examples of EMN's prototype-based explanations across diverse datasets, including MNIST~\cite{lecun2010mnist}, HAM10000~\cite{tschandl2018ham10000}, and CUB-200~\cite{Wah2011}. For each input query (Query A–D columns; gray background), EMN retrieves the top-3 most similar prototypes from memory, displayed in the corresponding Prototype A–D columns (white background).

The selected prototypes share not only the same class label but also fine-grained visual features with the query. For MNIST~\cite{lecun2010mnist} Query C, the retrieved prototypes match the query digit ``5'' in stroke style---specifically, the curvature of the upper loop and overall slant. For HAM10000~\cite{tschandl2018ham10000} Query D, the prototypes share diagnostically relevant features such as the central dark pigmentation pattern and irregular border shape. For CUB-200~\cite{Wah2011} Query A, all three prototypes depict birds clinging vertically to a tree trunk---matching not just the species but also the pose of the query image.

These observations qualitatively support the subclass alignment results in Section~\ref{sec:subclass}: EMN selects prototypes that capture instance-specific visual details beyond class-level similarity. This suggests that the memory-refined latent preserves fine-grained information, enabling more semantically meaningful explanations across domains.

\section{Conclusion}\label{sec:con}
Existing prototype-based models often exhibit dataset-dependent explanation quality, limiting their reliability across domains. 
We propose the EMN, which mitigates this limitation by grounding prototype selection in memory-refined latent representations that jointly support reconstruction and classification. 
EMN retrieves specific training instances as explanations; this instance-based retrieval resembles episodic memory retrieval in cognitive neuroscience~\cite{tulving1972episodic}, which involves recollecting specific events along with their contextual details.
The experimental results highlight the potential of incorporating brain-inspired memory dynamics into deep learning systems to improve explainability with a modest accuracy trade-off.
Limitations include the absence of formal convergence or stability analysis for the iterative update rule, reliance on proxy metrics rather than human evaluation studies, and baseline comparisons that follow each method's official protocol without unified reimplementation.
A promising direction is to extend EMN toward concept-level (more semantic) prototypes that bridge instance retrieval and higher-level abstraction; we leave this as future work.


\section*{Acknowledgment}
This paper was prepared with the assistance of Claude (Anthropic). The AI tool was used for writing assistance and code review.

\bibliographystyle{IEEEtran}
\bibliography{references}

\end{document}